\documentclass[letterpaper, 10 pt, conference]{ieeeconf}  

\IEEEoverridecommandlockouts                              

\usepackage[utf8]{inputenc} 
\usepackage[T1]{fontenc}    
\usepackage{url}            
\usepackage{booktabs}       
\usepackage{amsfonts}       
\usepackage{nicefrac}       
\usepackage{microtype}      
\usepackage{xcolor}         
\usepackage[ruled]{algorithm2e}
\SetAlFnt{\small}
\usepackage{amsmath}
\usepackage{bm}
\usepackage{bbold}
\usepackage{graphicx}
\usepackage{multirow}
\usepackage{textcomp}
\usepackage{wrapfig}
\usepackage{xspace}
\usepackage{interval}
\usepackage{bm,upgreek}
\usepackage{colortbl}
\let\labelindent\relax
\usepackage{enumitem}
\usepackage{amssymb}
\usepackage{pifont}
\usepackage{cite}

\definecolor{LightCyan}{rgb}{0.88,1,1}
\definecolor{DarkGreen}{rgb}{0.0, 0.5, 0.0}
\definecolor{PaleAqua}{rgb}{0.74, 0.83, 0.9}

\title{\LARGE \bf
Hierarchical Point Attention for Indoor 3D Object Detection
}

\author{Manli Shu$^{1,2,\ast}$ Le Xue$^{2}$ Ning Yu$^{2}$ Roberto Martín-Martín$^{2,3}$ \\ Caiming Xiong$^{2}$ Tom Goldstein$^{1}$ Juan Carlos Niebles$^{2,4}$ and Ran Xu$^{2}$ 
\thanks{$^{1}$ Department of Computer Science, University of Maryland, College Park, MD, U.S.A. {\tt\small \{manlis, tomg\}@umd.edu}}%
\thanks{$^{2}$ Salesforce Research, Palo Alto, CA, U.S.A. {\tt\small \{manli.shu, lxue, ning.yu, cxiong, ran.xu\}@salesforce.com}}%
\thanks{$^{3}$ Department of Computer Science at the University of Texas at Austin, TX, U.S.A. {\tt\small robertomm@cs.utexas.edu}}%
\thanks{$^{4}$ Department of Computer Science, Stanford University, CA, U.S.A. {\tt\small jniebles@cs.stanford.edu}}%
\thanks{$^{\ast}$ Work done while at University of Maryland.}%
}

\begin{document}

\maketitle
\thispagestyle{empty}
\pagestyle{empty}

\begin{abstract}

3D object detection is an essential vision technique for various robotic systems, such as augmented reality and domestic robots. Transformers as versatile network architectures have recently seen great success in 3D point cloud object detection. 
However, the lack of hierarchy in a plain transformer restrains its ability to learn features at different scales. Such limitation makes transformer detectors perform worse on smaller objects and affects their reliability in indoor environments where small objects are the majority.
This work proposes two novel attention operations as generic hierarchical designs for point-based transformer detectors. First, we propose Aggregated Multi-Scale Attention (MS-A) that builds multi-scale tokens from a single-scale input feature to enable more fine-grained feature learning. Second, we propose Size-Adaptive Local Attention (Local-A) with adaptive attention regions for localized feature aggregation within bounding box proposals. 
Both attention operations are model-agnostic network modules that can be plugged into existing point cloud transformers for end-to-end training. 
We evaluate our method on two widely used indoor detection benchmarks. By plugging our proposed modules into the state-of-the-art transformer-based 3D detectors, we improve the previous best results on both benchmarks, with more significant improvements on smaller objects.

\end{abstract}

\section{INTRODUCTION}
\label{sec:intro}

3D computer vision models (\textit{e.g.}, object detectors) help robotic and control systems perceive and understand the environment from 3D data (\textit{e.g.}, point cloud), which provides more accurate geometric and spatial information and is robust to illumination and domain shifts. 
Since point clouds do not have a grid-like structure as images, previous works have proposed various neural network architectures for point cloud understanding~\cite{Graham2018sparse, landrieu2018graph, qi2018frustum, pointnet, qi2016volumetric, pointnet2, pointrcnn, su2018splat, xu2018spidercnn, yang2018folding, guan2022m3detr, 3detr, groupfree}.
With the success of attention-based architectures (\textit{i.e.}, transformers) in other learning regime~\cite{vaswani2017attention,dosovitskiy_image_2021,liu_swin_2021}, it has recently been applied to point clouds~\cite{pointtx2021zhao, pointformer, zhang2022patch, yu2022pointbert, yang2019pat, liu2022transgraph, kini3dmodt2023}. 
Some properties of transformers make them ideal for modeling point clouds. For example, their permutation-invariant property is necessary for modeling unordered sets like point clouds, and their attention mechanism helps learn long-range relationships and capture global context. %


Despite the advantages of transformers for point clouds, we find that the state-of-the-art transformer detectors have imbalanced performance across different object sizes, with lower average precision on smaller objects (see Section~\ref{sec:size-analysis}). Such imbalanced performance can affect the robustness of downstream applications, especially for indoor scenarios where the environments are cluttered with small objects~\cite{scannetv2, sunrgbd}. 
We speculate such bias against small objects can be attributed to two factors. 
First, for efficiency, existing models are trained on downsampled point clouds with far fewer points than the raw data. The extensive downsampling loses geometric details and impacts more significantly on smaller objects. 
Second, plain transformers~\cite{vaswani2017attention,dosovitskiy_image_2021} only learn features at the global scale, whereas smaller objects may require more fine-grained feature extraction.

Motivated by the observations, we expect point cloud transformers to benefit from hierarchical feature learning strategies~\cite{feng2023hierarchical, wen2023pyramid}, \textit{e.g.}, multi-scale and localized feature learning. 
Nonetheless, considering the computation intensity of point cloud transformers, using higher-resolution (\textit{i.e.}, higher point density) point cloud features can be inefficient. Furthermore, due to the irregularity of point clouds, it is non-trivial to integrate hierarchical designs into transformers for point-based 3D object detection.

\textbf{Our approach.} We propose two point-based attention modules for hierarchical feature learning on point clouds. Both modules are model-agnostic and can be plugged into any existing point-based transformers for end-to-end training. 

We first propose \textit{Aggregated Multi-Scale Attention} (MS-A), which builds higher resolution features from a single-scale input. It then uses the multi-scale features for cross-attention via multi-scale token aggregation~\cite{shunted2022ren} with little parameter overhead. 
The second proposed module is \textit{Size-Adaptive Local Attention} (Local-A), where the attention regions are defined by the detector's box proposals for each object candidate. It thus allows adaptive local feature learning.

We evaluate our method on two widely used indoor 3D detection benchmarks, ScanNetV2~\cite{scannetv2} and SUN RGB-D~\cite{sunrgbd}. We plug our attention modules into existing transformer-based 3D detectors and perform end-to-end training. Our method improves the previous best results 
with little parameter overhead and with more significant improvements on smaller objects. 
We summarize our main contributions as follows: 
\vspace{-3pt}
\begin{itemize}[leftmargin=*]\setlength\itemsep{-1pt}
    \item We identify that point-based 3D transformer detectors have performance imbalance issues and perform worse on smaller objects. 
    \item Motivated by our observation, we propose two generic point attention operations that enable multi-scale and localized feature learning.
    \item We apply our method to various point-based transformer detectors and improve the previous best result on two widely used indoor 3D detection benchmarks.
\end{itemize}

\section{Related Works}

\noindent\textbf{Deep neural networks for 3D point cloud.}
Existing network architectures for point cloud learning can be roughly divided into two categories based on their point cloud representation: \textit{grid-based} and \textit{point-based}, yet in between, some hybrid architectures operate on both representations~\cite{zhou2018voxelnet, lang2019pointpillars, shi2020pvrcnn, ye2020hvnet, wang2022cagroup}. \textit{Grid-based} methods project the irregular point clouds into grid-like structures, such as 3D voxels or pillars~\cite{voxnet, riegler2017octnet, wang2017ocnn, he2022voxeltx, wang2023deeqnet}. 
\textit{Point-based} methods, on the other hand, directly learn features from the raw point cloud. Within this category, graph-based methods~\cite{simonovsky2017dynamic, wang2019graph, xu2020gcn, zhou2021graph} use graphs to model the relationships among the points. Other works regard the point cloud as a set and learn features through set abstraction~\cite{pointnet, pointnet2, ma2022pointmlp, wang2022rbgnet}. Recent works explore the transformer architecture for point-based learning~\cite{pointtx2021zhao, groupfree, 3detr, pointformer, zhang2022patch, wu2022ptxv2} by feeding individual points as tokens into a transformer, where the attention mechanism learns point features at a global scale. 
While most previous methods improve point cloud learning by developing new backbones, our work aims to provide a generic model-agnostic solution. PAConv~\cite{paconv2021xu} proposes a generic convolution operation while we study cross-attention operations that can be integrated into the emerging transformer models.

\noindent\textbf{Hierarchical designs for 3D vision transformers.}
Inspired by the literature in 2D vision~\cite{liu_swin_2021, shunted2022ren, li2022mvit,deformdetr} especially convolutional neural networks~\cite{lecun1989cnn} hierarchical designs for 3D transformers also seek to learn features at different granularities. 
Some methods~\cite{zhang2022patch,chen2022deformable} propose attention mechanisms that can do multi-scale feature learning, but they only work on voxels and cannot be applied to point-based 3D models. Lai et al.~\cite{stratified2022lai} proposed stratified self-attention that can learn multi-scale features directly on point clouds, but it only supports downsampling, which limits more fine-grained feature learning. Our multi-scale attention, on the other hand, can produce features of arbitrary resolutions.
Another line of work~\cite{pointtx2021zhao,pointformer,3detr} proposes to learn localized features by applying self-attentions to local regions specified by $k$ nearest neighbors or a fixed radius/window. Instead of a fixed region, our localized attention is size-adaptive, which uses adaptive local regions that match the approximated object sizes.

\begin{figure}[t]
  \centering
    \resizebox{\linewidth}{!}{
        \includegraphics{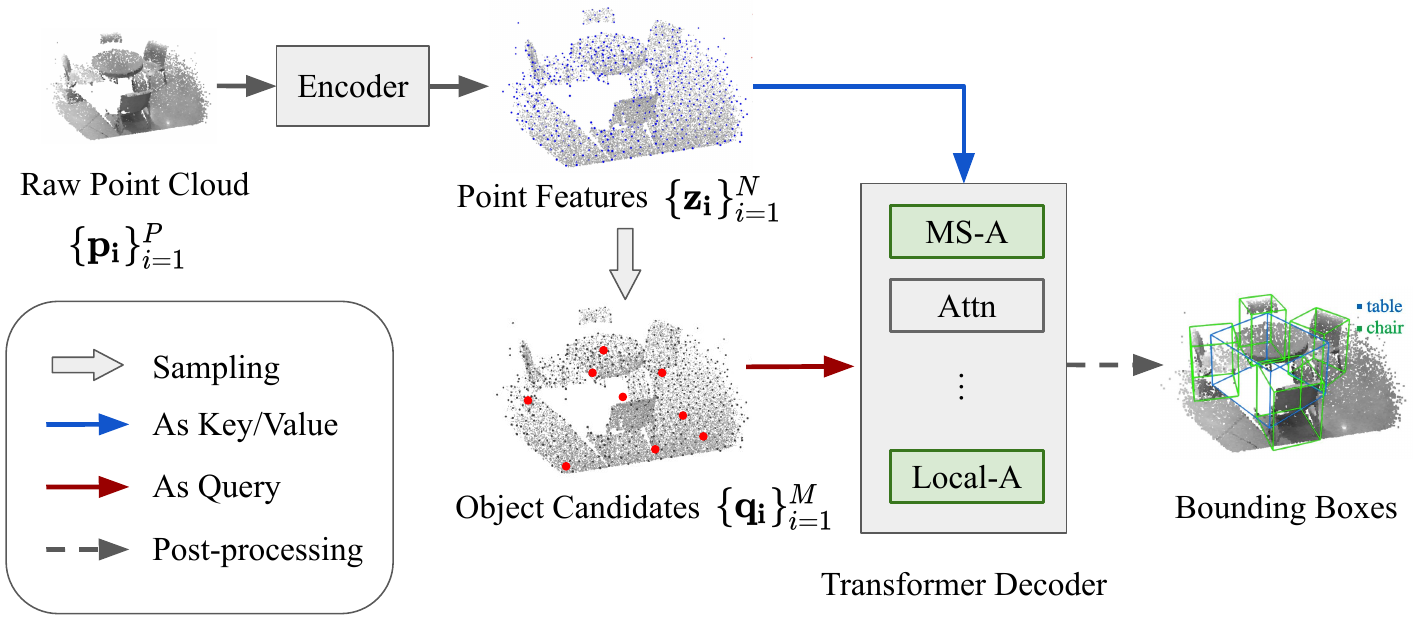}
        }
  \caption{\textbf{A point-based 3D transformer detectors with our proposed modules (MS-A and Local-A).} Detector overview: the raw point cloud is downsampled during encoding to obtain the point features, which serve as the key and value in the transformed decoder. Object candidates are sampled from the point features (e.g., using FPS). The transformer decoder learns object features via alternating self- and cross-attentions. The proposed MS-A and Local-A are cross-attention modules that can be plugged into the transformer. See Fig.~\ref{fig:ms-a} and Fig.~\ref{fig:local-a} for the design details of each module.} 
  \label{fig:intro}
\end{figure}

\section{Method}
\label{sec:method}

\subsection{Background}
\noindent\textbf{Point cloud object detection.} Given a point cloud $\mathcal{P}_\text{raw}$ with a set of $P$ points $\mathcal{P} = \{\mathbf{p_i}\}_{i=1}^{P}$, each point $p_i \in \mathbb{R}^{3}$ is represented by its 3-dimensional coordinate. 3D object detection on point cloud aims to predict a set of bounding boxes for the objects in the scene, including their locations (as the center of the bounding box), size and orientation of the bounding box, and the semantic class of the corresponding object. Note that due to the computation limit, the point cloud is downsampled at the early stage of a model to a subset of $\mathcal{P}_\text{raw}$, which contains $N$ ($N << P$) points. $\mathcal{P} = \text{SA}(\mathcal{P}_\text{raw}) = \{\mathbf{p_i}\}_{i=1}^{N}$ contains the aggregated groups of points around $N$ group centers, where SA (set abstraction) is the aggregation function, and the group centers are sampled from the raw point cloud using \textit{Farthest Point Sample (FPS)}~\cite{pointnet}, a random sampling algorithm that provides good coverage of the entire point cloud.

\noindent\textbf{Point-based 3D transformer detectors.} Our method is built on point-based 3D object detectors~\cite{pointrcnn, votenet, groupfree}, which detect 3D objects in point clouds in a bottom-up manner. Compared to other 3D detectors that generate box proposals in a top-down manner on the bird's-eye view or voxelized point clouds~\cite{chen2017mv3d, TianYZLH23}, point-based methods work directly on the irregular point cloud without quantization errors. 

We illustrate the general point-based 3D transformer detector in Fig.~\ref{fig:intro}. The features of the input point cloud $\{\mathbf{z_i}\}_{i=1}^{N}, \mathbf{z_i} \in \mathbb{R}^{d}$ is obtained using a backbone model (\textit{e.g.}, PointNet++~\cite{pointnet2}), where $d$ is the feature dimension. 
Point-based detectors generate bounding box predictions starting with $M$ ($M < N$) initial object \textit{candidates} $\{\mathbf{q_i}\}_{i=1}^{M}, \mathbf{q_i} \in \mathbb{R}^{C}$, sampled from the point cloud as object centers. A common candidate sampling approach is the Farthest Point Sample (FPS). Once the initial candidates are obtained, the detector extracts features for every object candidate. Attention-based methods~\cite{groupfree} learn features by doing self-attention among the object candidates and cross-attention between the candidates (\textit{i.e.}, query) and point features $\{\mathbf{z_i}\}_{i=1}^{N}$. 

The learned features of the object candidates will then be passed to prediction heads, which predict the attributes of the bounding box for each object candidate. The attributes of a 3D bounding box include its location (box center) $\hat{\mathbf{c}} \in \mathbb{R}^3$, size $\hat{\mathbf{d}}\in \mathbb{R}^3$, orientation (heading angles) $\hat{\mathbf{a}}\in \mathbb{R}$, and the semantic label of the object $\hat{\mathbf{s}}$. With these parameterizations, we can represent a bounding box proposal as $\mathbf{\hat{b}} = \{\hat{\mathbf{c}}, \hat{\mathbf{d}}, \hat{\mathbf{a}}, \hat{\mathbf{s}}\}$. 

\noindent\textbf{Attention mechanism} is the basic building block of transformers. The attention function takes in query ($Q$), key ($K$), and value ($V$) as the input. The output of the attention function is a weighted sum of the value, with the attention weight being the scaled dot-product between the key and query:
\begin{align}
\text{Attn}(Q, K, V) = \text{softmax}(\frac{QK^T}{\sqrt{d_h}})V,
\end{align}
where $d_h$ is the hidden dimension of the attention layer.
For self-attention, $Q \in \mathbb{R}^{d_h}$, $K \in \mathbb{R}^{d_h}$ and $V \in \mathbb{R}^{d_v}$ are transformed from the input $X \in \mathbb{R}^{d}$ via linear projection with parameter matrix $W_{i}^{Q} \in \mathbb{R}^{d \times d_h}$, $W_{i}^{K} \in \mathbb{R}^{d \times d_h}$, and $W_{i}^{V} \in \mathbb{R}^{d \times d_v}$ respectively. For cross-attention, $Q$, $K$, and $V$ can have different sources.

In practice, transformers adopt the \textbf{multi-head attention} design, where multiple attention functions are applied in parallel across different attention \textit{head}s. The input of each attention head is a segment of the layer's input. Specifically, the query, key, and value are split along the hidden dimension into $(Q_i, K_i, V_i)_{i=1}^{h}$, with $Q_i \in \mathbb{R}^{d_h/h}, K_i \in \mathbb{R}^{d_h/h}, V_i \in \mathbb{R}^{d_v/h}$, where $h$ is the number of attention heads. The final output of the multi-head attention layer is the projection of the concatenated outputs of all attention heads:
\begin{align}
\begin{split}
\text{MultiHead}(Q, & K, V) = \text{Concat}(\{\text{Attn}(Q_0, K_0, V_0);  \\
                            & ...; \text{Attn}(Q_{h-1}, K_{h-1}, V_{h-1})\})W^{O},
\end{split}
\end{align}
where the first term denotes the concatenation of the output and $W^{O}$ is the output projection matrix.

\subsection{Aggregated Multi-Scale Attention} 

In existing point-based transformer detectors, the cross-attention modules are applied between object candidates and all other points of the point cloud. 
However, due to the computation overhead of the attention function, the actual point cloud that the model uses is a downsampled set of 1024 points~\cite{groupfree, 3detr}, whereas the raw point cloud usually contains tens of thousands points~\cite{sunrgbd, scannetv2}. Such extensive downsampling causes a loss of detailed geometric information and fine-grained features essential for dense prediction tasks like object detection.

To this end, we propose \textit{Aggregated Multi-Scale Attention} (MS-A), which learns to build arbitrary higher-resolution (\textit{i.e.}, higher point density) feature maps from the single-scale feature input. It then uses features of both features as the multi-scale key/value in the cross-attention between object candidates and other points. The multi-scale feature aggregation is realized via multi-head token aggregation, where we use the key and value of different scales in different subsets of attention heads. We aim to provide fine-grained geometric details for object-level feature learning. %

\begin{figure}[t]
  \centering
    \resizebox{0.95\linewidth}{!}{
        \includegraphics{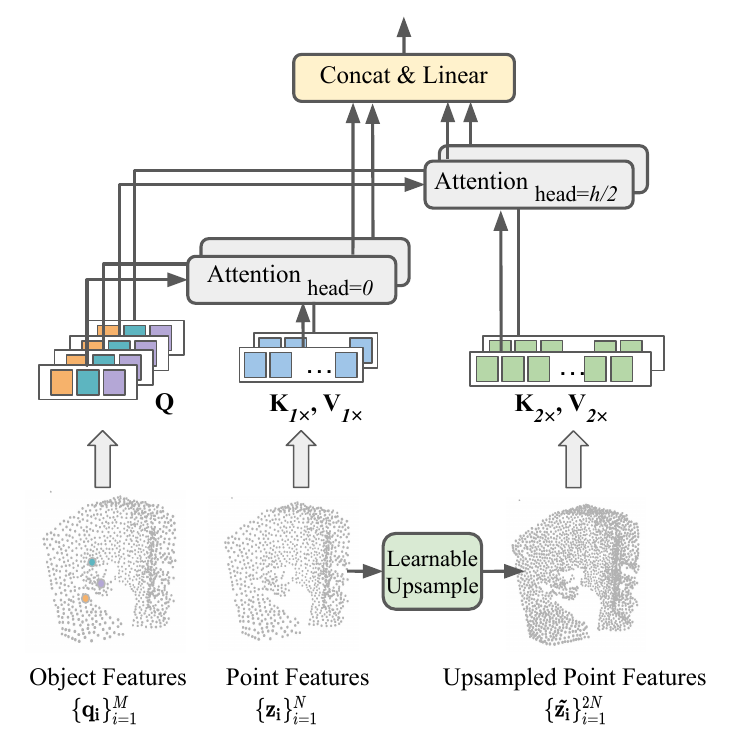}
        }
  \caption{\textbf{Aggregated Multi-Scale Attention (MS-A)} learns features at different scales within the multi-head cross-attention design. It constructs higher resolution (\textit{i.e.}, higher point density) point features from the single-scale input point features and uses keys and values of both scales.}
  \label{fig:ms-a}
\end{figure}

The first step of MS-A is to build a higher-resolution feature map from the single-scale input. We propose a learnable upsampling procedure. Given the layer's input point cloud feature $\{\mathbf{z_i}\}_{i=1}^{N}, \mathbf{z_i} \in \mathbb{R}^{d}$, MS-A can create a feature map with an arbitrary number ($N'$) of points. 
To get the locations (\textit{i.e.}, coordinates) of the $N'$ points, we use FPS to sample $N'$ points from the \textit{raw point cloud} and obtain $\{\mathbf{p_i}\}_{i=1}^{N'}, \mathbf{p_i} \in \mathbb{R}^{3}$. 
Next, for each sampled point $\mathbf{p_i}$ in the $N'$-point cloud, we initialize their feature via three-nearest-neighbor weighted interpolation~\cite{pointnet2}, a generic operation adopted by many point-based methods~\cite{votenet, pointtx2021zhao, paconv2021xu}. Unlike previous work, which uses the interpolation for the skip connection between the encoder and decoder, our $N'$ points are sampled arbitrarily from the raw point cloud. We can sample a $N'$-point feature to have more points than any feature map in the model, thus providing more fine-grained geometry details.

From the interpolated feature initialization, we obtain the final feature representation of the $N'$-point feature map via a learnable module $\Phi_{\theta}$. We use MLP as the learnable projection function. The learned $N'$-point feature map is: 
\begin{align}
    & \{\mathbf{\tilde{z_i}}\}_{i=1}^{N'}, \, \mathbf{\tilde{z_i}} = \Phi_{\theta}(\text{interpolate}(\{z_{i}^{0}, z_{i}^{1}, z_{i}^{2}\}))
\end{align}
After the upsampling, we have two sets of point features of different scale $\{\mathbf{z_i}\}_{i=1}^{N}, \{\mathbf{\tilde{z_i}}\}_{i=1}^{N'}$. To avoid computation increase, we perform multi-head cross-attention on both sets of point features in a single pass. We use features of different scales on different attention heads. We divide attention heads evenly into two groups and use $\{\mathbf{z_i}\}_{i=1}^{N}$ to derive $K$ and $V$ in the first group while using $\{\mathbf{z_i}\}_{i=1}^{N'}$ for the other. Both groups share the same set of queries derived from object candidates $\{\mathbf{q_i}\}_{i=1}^{M}$. 
Since the input and output of this module are the same as a plain attention module, we can plug MS-A into any attention-based model to enable feature learning at different scales. In practice, we set $N'=2N$ and apply MS-A only at the first layer of a transformer decoder to introduce minimal computation overhead. Further analyses on design choices are in Section~\ref{sec:ablation}.

\subsection{Size-Adaptive Local Attention} 
\label{sec:method-local}

Although the attention mechanism can effectively model the long-range relationship between points, it cannot guarantee that the learned module will pay more attention to points that are important to a particular object (\textit{e.g.}, those belonging to the object). 
On the other hand, the lack of hierarchy in plain transformers does not support explicit localized feature extraction. 
While some work addresses this issue by restricting attention to fixed local regions~\cite{3detr, zhang2022patch}, we propose \textit{Size-Adaptive Local Attention} (Local-A) that defines local regions adaptively based on the size of each bounding box proposal. 

Point-based 3D detectors can generate intermediate bounding box proposals $\{\mathbf{\hat{b}_i}\}_{i=1}^{M}$ from the object features ($\{\mathbf{q_i}\}_{i=1}^{M}$) of a decoder layer. Given the bounding boxes generated for each object candidate $\mathbf{q_i}$, we perform cross-attention between $\mathbf{q_i}$ and points sampled from within its corresponding box $\mathbf{\hat{b}_i}$. We thus have customized size-adaptive attention regions for every query point.
For every input object candidate $\mathbf{q_i}^{l} \in \mathbb{R}^{d}$ at decoder layer $l$, it is updated by Local-A via ($i$ denotes the index of a query point):
\begin{align}
   & \mathbf{q_i}^{l+1} = \text{Attn}(Q_i^{l}, K_i, V_i), \text{where } \\ 
   & Q_i^{l} = \mathbf{q_i}^{l} W^{Q}, K_i = Z_i W^{K},  V_i = Z_i W^{V} \text{with } \\ 
   & Z_i = \{\mathbf{z_j}^{i} \,|\, \mathtt{pos}(\mathbf{z_j}^{i}) \text{ in } \mathbf{\hat{b}_i}\} , \mathbf{\hat{b}_i} = \text{Pred}^{l}_{box}(\mathbf{q_i}^{l}). \label{eq:boxquery}
\end{align}

\begin{figure}[t]
  \centering
    \resizebox{0.95\linewidth}{!}{
        \includegraphics{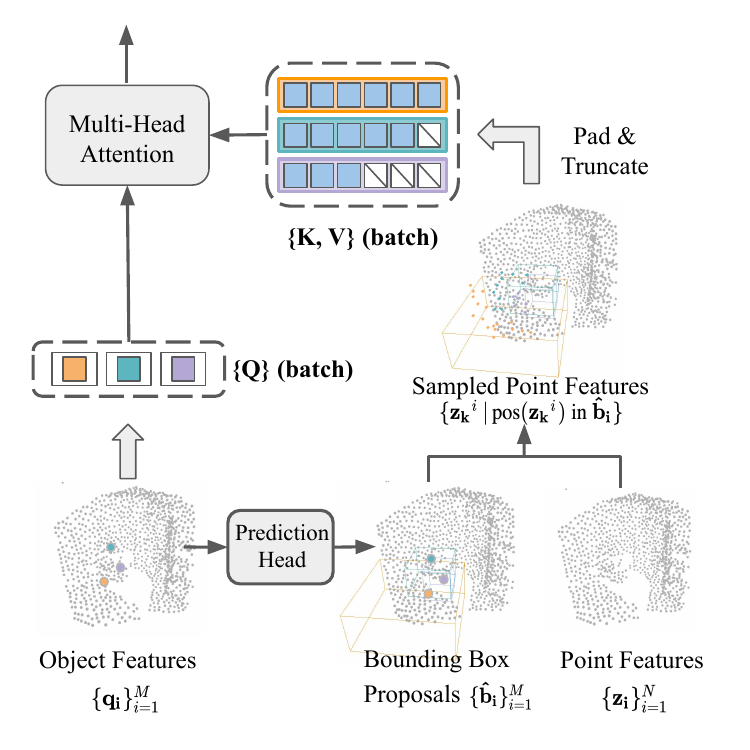}
        }
  \caption{\textbf{Size-Adaptive Local Attention (Local-A)} performs adaptive local attention between each object candidate (query) and the points inside its corresponding bounding box proposal. The attention range (the token lengths of keys and values) varies across different object candidates, so perform padding/truncating to allow batch processing.} 
  \label{fig:local-a}
\end{figure}

In Eq(\ref{eq:boxquery}), $\mathtt{pos}(\cdot)$ obtains the 3D-coordinate $(x, y, z)$ of a point feature $\mathbf{z}$, and $\mathbf{\hat{b}_i}$ is the model's bounding box prediction for query point $\mathbf{q_i}^{l}$ (generated via the prediction head $\text{Pred}^{l}_{box}$). $Z_i$ hereby denotes a set of points \textit{inside} box $\mathbf{\hat{b}_i}$. $K_i$ and $V_i$ are derived from this set of point features. 
Note that $\{\mathbf{z_i}\}_{i=1}^{N}$ is the entire set of point features extracted by the encoder and is not updated during decoding.  

Since each $\mathbf{q_i}$ has its own sets of keys and values depending on the size of its bounding box predictions, the size of $K_i$/$V_i$ differs across queries. To allow batch computation, we set a maximum number of points ($N_{local}$) for the sampling process and use $N_{local}$ as a fixed token length for every query point. For bounding boxes that contain less than $N_{local}$ points, we pad the point sequence with an unused token to $N_{local}$ and mask the unused tokens in the cross-attention function; for those containing more than $N_{local}$ points, we randomly discard them and truncate the sequence to have $N_{local}$ points as keys and values. Lastly, when the bounding box is empty, we perform ball query~\cite{pointnet} around the object candidate to sample $N_{local}$ points. 

Like MS-A, Local-A does not pose additional requirements on a module's input. Therefore, we can apply it to any transformer layer. We apply Local-A at the end of a transformer where bounding box proposals are generally more accurate. We empirically set $N_{local}=16$ with related ablation study in Section~\ref{sec:ablation}.

\section{Experiments}
In this section, we first evaluate our method on two widely used indoor point cloud detection datasets, ScanNetV2 and SUN RGB-D. Next, we provide qualitative analyses of visualizations of bounding boxes and attention weights. We also conduct evaluations with our proposed size-aware metrics. Lastly, we include ablation studies on the design choices of our method. 

\noindent\textbf{Datasets.} 
\textit{ScanNetV2}~\cite{scannetv2} consists of 1513 reconstructed meshes of hundreds of indoor scenes, with rich annotations for 3D scene understanding tasks, including classification, semantic segmentation, and object detection. For object detection, it provides axis-aligned bounding boxes with 18 object categories. We follow the official dataset split by using 1201 samples for training and 312 samples for testing. 
\textit{SUN RGB-D}~\cite{sunrgbd} is a single-view RGB-D dataset with 10335 samples. For 3D object detection, it provides oriented bounding box annotations with 37 object categories, while we follow the standard evaluation protocol~\cite{votenet} and only use the ten common categories. The training and testing split contains 5285 and 5050 samples, respectively. 

\noindent\textbf{Evaluation metrics.} 
For both datasets, we follow the standard evaluation protocol~\cite{votenet} and use the mean Average Precision (mAP) as the evaluation metric. We report mAP scores under two different Intersection over Union (IoU) thresholds: mAP@0.25 and mAP@0.5. In addition, in Section~\ref{sec:size-analysis}, to evaluate model performance across different object sizes, we follow the practice in 2D vision~\cite{coco} and implement our size-aware metrics that measure the mAP on small, medium, and large objects, respectively.
Because of the randomness of point cloud training and inference, we train a model 5 times and test each model 5 times. We report both the best and the average results among the 25 trials. 

\noindent\textbf{Baselines.} 
We validate our method and demonstrate its model-agnostic property by applying it to three different transformer-based point cloud detectors:      
Group-Free~\cite{groupfree}, RepSurf~\cite{ran2022surface}, and 3DETR~\cite{3detr}. Group-Free encodes point features with a PointNet++~\cite{pointnet2} backbone and learns object features using a transformer decoder with plain attention. 
We consider two Group-Free configurations: Group-Free$^{6, 256}$ and Group-Free$^{12, 512}$, where Group-Free$^{L,O}$ denotes the variant with $L$ decoder layers and $O$ object candidates.   
RepSurf-U learns point features from a novel multi-surface representation explicitly describing local geometry. It uses a similar transformer decoder as \cite{groupfree} and has two configurations. The official implementation and the averaged results of RepSurf-U for object detection are not publicly available, so we include the results of our reproduction of RepSurf-U.     
3DETR is another end-to-end 3D transformer detector. Unlike other baselines, 3DETR's encoder and decoder are both transformers. The official result of 3DETR is not over multiple runs, so we report our reproduced results for this baseline.

We also include previous point-based 3D detectors for comparison. VoteNet~\cite{votenet} aggregates features for object candidates through end-to-end optimizable Hough Voting. H3DNet~\cite{h3dnet} proposes a hybrid set of geometric primitives for object detection and trains multiple individual backbones for each primitive. Pointformer~\cite{pointformer} proposes a hierarchical architecture as the backbone and adopts the voting algorithm of VoteNet for object detection.

\noindent\textbf{Implementation details.} 
For a baseline model with $L$ transformer layers, we enable multi-scale feature learning by replacing the cross-attention of the $1$-st layer with MS-A. At the $L$-th layer (\textit{i.e.}, the last decoder layer), we replace its cross-attention with Local-A. We follow the original training settings of the baseline models~\cite{groupfree, ran2022surface, 3detr}. 

\begin{table}[ht]
\caption{\textbf{Performance of object detection on ScanNetV2.} }
\vspace{-1em}
\footnotesize {We follow the standard protocol~\cite{votenet} by reporting the best results over 25 trials (5 trainings, each with 5 testings) with the averaged results in the bracket. Backbone stands for the encoder model architecture, and PN++ denotes PointNet++.
Group-Free$^{L, O}$ denotes the variant with $L$ decoder layers and $O$ object candidates. The same notation applies to RepSurf-U. 
Note that the detection code of RepSurf is not published, so we implement our version of RepSurf-U and report the reproduced results (repd.). }
\begin{center}
\scalebox{0.82}{
\begin{tabular}{c|c|c|cc}
\toprule
\multirow{2}{*}{Methods}    & \multirow{2}{*}{\#Params} & \multirow{2}{*}{Backbone}  & \multicolumn{2}{c}{ScanNet V2}  \\
\cmidrule(lr){4-5} 
                            & &   & mAP@0.25 & mAP@0.50  \\
\midrule
VoteNet~\cite{votenet}    &   -    &  PN++ &  62.9  &  39.9  \\
H3DNet~\cite{h3dnet}     &   -    &  PN++ &  64.4   & 43.4 \\
H3DNet~\cite{h3dnet}     &   -    &  4$\times$PN++ & 67.2   & 48.1   \\
Pointformer~\cite{pointformer}  &   -    &  transformer &  64.1    &  42.6   \\
\midrule 
3DETR-m~\cite{3detr} (repd.)  &   7.4M    &  transformer &  64.1 (62.9)    &  45.8 (44.1)   \\
\rowcolor{white} w/ MS + Local (Ours)      &   7.5M    &  transformer &   \textbf{64.8 (63.7)}  &  \textbf{46.4 (45.2)}   \\
\midrule 
Group-Free$^{6, 256}$~\cite{groupfree} &  14.5M  &  PN++ &  67.3 (66.3)   &  48.9 (48.5)  \\
\rowcolor{white} w/ MS + Local (Ours)  &  14.6M  &  PN++  &    67.8 (66.8)  &    50.8 (49.5)    \\
RepSurf-U$^{6, 256}$~\cite{ran2022surface}  &  14.5M   &  PN++ &  68.8 (  ~~-~  )  &  50.5 (  ~~-~  )   \\
RepSurf-U$^{6, 256}$ (repd.)  &  14.5M   &  PN++ &  68.0 (67.4)   &  50.2 (48.7)  \\
\rowcolor{white} w/ MS + Local (Ours)  &  14.6M  &  PN++   &    \textbf{69.9 (68.7)}  &    \textbf{53.0 (51.3)}  \\
\midrule
Group-Free$^{12, 512}$~\cite{groupfree} &  29.6M   &  PN++w2x &  69.1 (68.6)   & 52.8 (51.8) \\
\rowcolor{white} w/ MS + Local (Ours)  &  29.6M &  PN++w2x &    70.3 (69.0)   &    53.5 (52.3)  \\
RepSurf-U$^{12, 512}$~\cite{ran2022surface}  &  29.7M   &  PN++w2x & 71.2 (  ~~-~  )   & 54.8 (  ~~-~  )  \\
RepSurf-U$^{12, 512}$ (repd.)  &  29.7M   &  PN++w2x & 70.8 (70.2)   & 54.4 (53.6)  \\
\rowcolor{white} w/ MS + Local (Ours)  &  29.8M &  PN++w2x &    \textbf{71.4 (70.6)}   &    \textbf{55.6 (54.3)}   \\

\bottomrule
\end{tabular}
}
\end{center}
\vspace{-1em}

\label{tab:scannet}
\end{table}
\begin{table}[ht]
\caption{\textbf{Performance of object detection on SUN RGB-D.}}
\vspace{-1em}
\begin{center}
\scalebox{0.85}{
\begin{tabular}{c|cc}
\toprule
Methods    & mAP@0.25 & mAP@0.50  \\
\midrule
VoteNet~\cite{votenet}     & 59.1 &  35.8 \\
H3DNet~\cite{h3dnet}      & 60.1 & 39.0 \\
3DETR-m~\cite{3detr}     &  59.1  &  32.7  \\
Pointformer~\cite{pointformer}  & 61.1  &  36.6  \\
\midrule 
Group-Free$^{6, 256}$~\cite{groupfree} &   63.0 (62.6)  &  45.2 (44.4)  \\
\rowcolor{white} w/ MS + Local (Ours)  &   63.3 (62.8)   &  45.8 (44.9)  \\
RepSurf-U$^{6, 256}$~\cite{ran2022surface}  &   64.3 (  ~~-~  )  &  45.9 (  ~~-~  )   \\
RepSurf-U$^{6, 256}$ (repd.)  & 64.0 (63.3)  &  45.7 (45.2)  \\
\rowcolor{white} w/ MS + Local (Ours) &  \textbf{64.2} (\textbf{63.6})    &   \textbf{47.0} (\textbf{45.7})  \\
\bottomrule
\end{tabular}
}
\end{center}

\label{tab:sunrgbd}
\end{table}
\begin{figure}[!ht]
  \centering
    \resizebox{\linewidth}{!}{
        \includegraphics{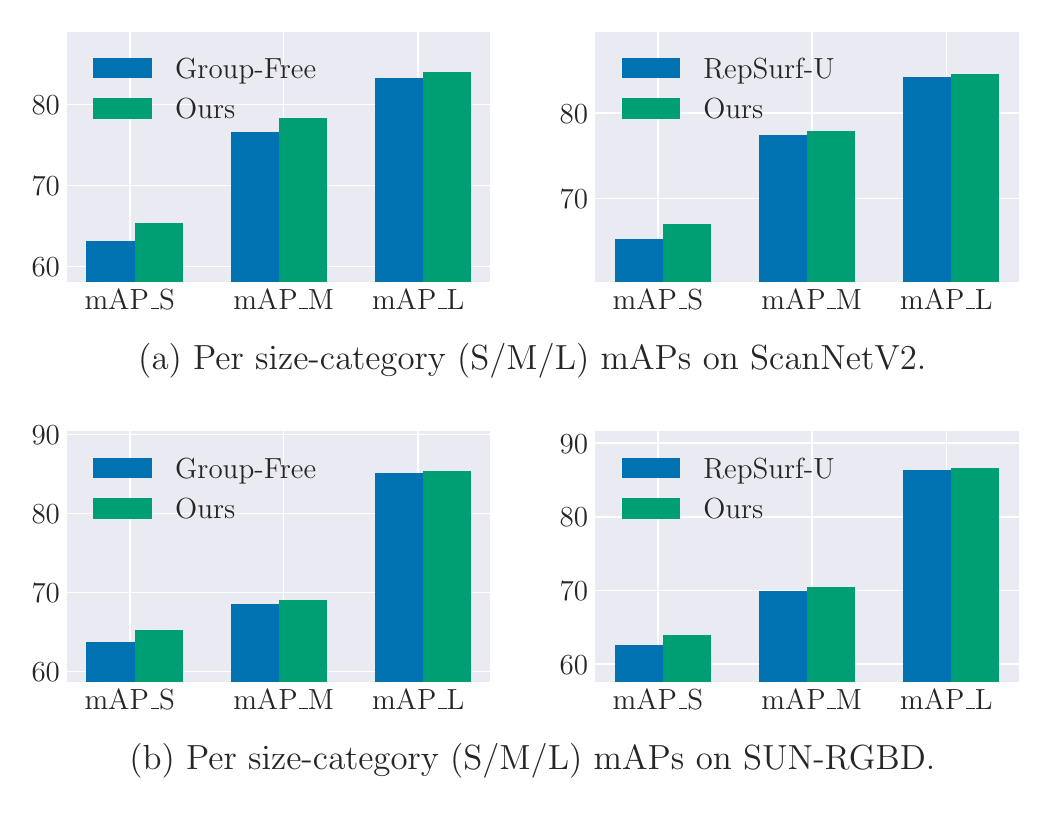}
        }
  \caption{\textbf{Performance on object of different sizes}. We define the S/M/L thresholds based on each dataset's statistics (volume distribution).}
  \label{fig:size-eval}
\end{figure}

\subsection{Main Results}
In Table~\ref{tab:scannet}, we observe consistent improvements in baseline models when equipped with our attention modules. On ScanNetV2, we perform on par with the state-of-the-art RepSurf-U detector by applying MS-A and Local-A to Group-Free, and we can further improve RepSurf-U across varying model configurations with little parameter overhead. 
Table~\ref{tab:sunrgbd} shows a similar trend on SUN RGB-D. Our method has larger improvements on ScanNetV2 than SUN RGB-D, which can be attributed to the different complexity of the two datasets: a ScanNetV2 sample contains the entire scan of a scene, whereas a SUN RGB-D sample only contains a single view. The benefit of our hierarchical attention is more prominent when handling complex scenes with fine-grained raw data. 

\begin{figure*}[!ht]
  \centering
    \resizebox{0.8\linewidth}{!}{
        \includegraphics{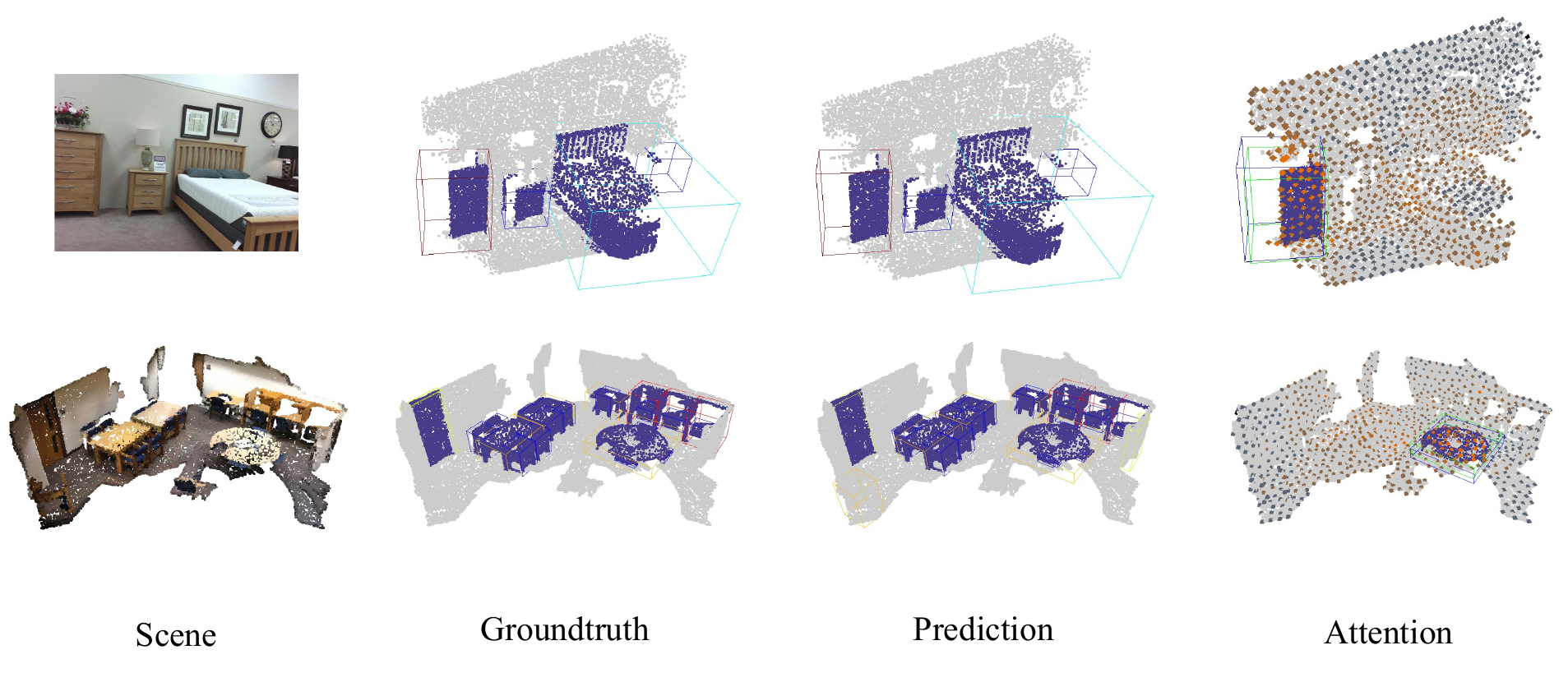}
        }
  \caption{\textbf{Qualitative results on SUN RGB-D (top) and ScanNetV2 (bottom)}. The color of a bounding box in the middle two columns stands for the semantic label of the object. In the last column, we draw both the ground truth (in green) and the prediction (in blue) of the object. We highlight the points that belong to an object for better visualization. In the last column, we visualize the attention weight of the last transformer layer (before applying Local-A). We visualize the cross-attention weight between an object candidate and the point cloud. }
  \label{fig:visualize}
\end{figure*}

\subsection{Performance on objects of different sizes.}
\label{sec:size-analysis}
In addition to the standard evaluation metrics, we also examine the models' performance across different object sizes. We define the size-aware metric for 3D detection following the tradition in 2D detection~\cite{coco}. We set the threshold for mAP$_S$ as the 30$th$ percentile of the volume of all objects and use the 70$th$ percentile as the threshold for mAP$_L$. 

We evaluate Group-Free and RepSurf models on both benchmark datasets. From the results in Fig.~\ref{fig:size-eval}, we can see that mAPs on smaller objects (mAP$_S$) are much lower than on larger objects for both models. When applied with our proposed attention modules, we observe the largest performance gains in mAP$_S$ . 

\subsection{Qualitative Results}

In Figure~\ref{fig:visualize}, we provide qualitative results on both datasets. The visualized results are of our methods applied to the Group-Free detectors. The qualitative results suggest that our model can detect and classify objects of different scales even in complex scenarios containing more than ten objects (\textit{e.g.}, the example in the bottom row). 
By looking into cross-attention weights in the transformer detector, we find that object candidates tend to have higher correlations with points that belong to their corresponding objects.

\subsection{Ablation Study}
\label{sec:ablation}


\noindent\textbf{The maximum number of points ($N_{local}$) in Local-A.} 
In Local-A, for each object candidate (i.e., query), we sample a set of points within its corresponding bounding box proposal and use the point features as the key and value for this object candidate in the cross-attention function. As introduced in Section~\ref{sec:method-local}, we cap the number of sampled points with $N_{local}$ to allow batch computation. 

\begin{table}[!ht]
\caption{\textbf{The effect of $N_{local}$ in Local-A.}}
\vspace{-1em}
\footnotesize {When there are enough points, a larger $N_local$ means the points are sampled more densely within each bounding box proposal.}
\begin{center}
\scalebox{0.8}{
\begin{tabular}{c|cc|ccc}
\toprule
 $N_{local}$  & mAP@0.25 & mAP@0.50 & mAP$_S$ & mAP$_M$ & mAP$_L$ \\
 \midrule
  8    &    67.8   &   51.1  & 64.3 & 77.2 & 82.8 \\
 16    &    \textbf{68.8}   &   \textbf{52.3}  & 65.1 & \textbf{77.9} & 83.4 \\
 24    &    68.7   &   \textbf{52.3}  & \textbf{65.2} & 77.7 & 83.5 \\
 32    &    68.3   &   52.1  & 64.7 & 77.3 & \textbf{83.8}\\
 \bottomrule
\end{tabular}
}
\end{center}
\vspace{-5pt}
\label{tab:ablate-refnp}
\end{table}

From Table~\ref{tab:ablate-refnp}, we find that too little number of points (e.g., $N_{local}=8$) for Local-A results in a performance drop. As $N_{local}$ increases, we do not observe a significant performance gain when it exceeds $N_{local}=16$. 
Intuitively, a small $N_{local}$ means the points within each bounding box are sampled sparsely, which can be too sparse to provide enough information about any object. This explains why $N_{local}=8$ does not work well. On the other hand, a large $N_{local}$ may only benefit large objects and have little effect on smaller objects because the latter are padded with unused tokens. 

\noindent\textbf{MS-A with different feature resolutions.} 
Similar to how the learnable upsampling in MS-A produces higher-resolution features, we can implement learnable \textit{downsampling} using conventional set abstraction~\cite{pointnet}, which aggregates point features within local groups and produce feature maps with fewer points (\textit{i.e.}, lower resolution). 
Intuitively, a higher-resolution feature map provides more fine-grained geometry details, while a more coarse one may provide a more global context. We conduct an empirical analysis on MS-A with different sampling ratios to study the effects of feature maps of different granularity. We choose a sampling ratio of 0.5 and 2.0 to represent coarse and fine-grained features, respectively.
\begin{table}[!ht]
\caption{\textbf{MS-A with different feature scales.}}
\vspace{-1em}
\footnotesize {Feature scale $= s$ means the feature map contains $s\times N$ points. A larger $s$ denotes a feature map with higher point density (\textit{i.e.}, resolution)}
\begin{center}
\scalebox{0.8}{
\begin{tabular}{l|cc|ccc}
\toprule
 Scales $s$  & mAP@0.25 & mAP@0.50 & mAP$_S$ & mAP$_M$ & mAP$_L$ \\
 \midrule
 $[1]$  &  68.6  & 51.8 & 63.1 & 76.6 & 83.2 \\
 $[1, 2]$  &  \textbf{68.9}  &  \textbf{52.5} & \textbf{65.0} & \textbf{77.5} & \textbf{83.9} \\
 $[0.5, 1, 2]$  &   67.9     &  51.7 &  64.6 & 76.7 & \textbf{83.9} \\
 \bottomrule
\end{tabular}
}
\end{center}
\label{tab:ablate-ms-ratio}
\end{table}

Results in Table~\ref{tab:ablate-ms-ratio} 
suggest that coarse features ($s = 0.5$) do not benefit transformer detectors. This is expected because transformers do not have limited receptive fields and thus do not need coarse-grained feature maps to learn global context. 

\section{CONCLUSION}

This work presents Aggregated Multi-Scale Attention (MS-A) and Size-Adaptive Local Attention (Local-A), two generic point-based attention operations that can be applied to various 3D transformer detectors and enable fine-grained feature learning. We improve point-based transformer detectors on two challenging indoor 3D detection benchmarks, with the largest improvement margin on smaller objects. 
As our method promotes fine-grained 3D feature learning, which is important to many 3D vision systems, future work will adapt the proposed attention modules to other applications, such as segmentation. 
Another future direction is to apply our method to improve outdoor detectors for applications like autonomous vehicles. Considering that mainstream outdoor detectors are usually not point-based, one may need to adapt the attention operations to other 3D representations.







\bibliographystyle{IEEEtran}
\bibliography{referece}

\end{document}